%% file: aaai25.tex
\documentclass[letterpaper]{article} 
\usepackage{aaai25}  
\usepackage{times}  
\usepackage{helvet}  
\usepackage{courier}  
\usepackage[hyphens]{url}  
\usepackage{graphicx} 
\usepackage{natbib}  
\usepackage{caption} 
\usepackage{algorithm}
\usepackage[noend]{algorithmic}

\usepackage{newfloat}
\usepackage{listings}
\DeclareCaptionStyle{ruled}{labelfont=normalfont,labelsep=colon,strut=off} 
\floatstyle{ruled}
\newfloat{listing}{tb}{lst}{}
\floatname{listing}{Listing}
\usepackage{amsmath}
\usepackage{amssymb}
\usepackage{amsthm}
\usepackage{cleveref}
\usepackage{subcaption}
\usepackage{comment}
\usepackage{multirow}
\usepackage{makecell}
\usepackage{array}

\DeclareMathOperator{\argmax}{arg\,max}
\DeclareMathOperator{\argmin}{arg\,min}

\DeclareMathOperator{\proj}{proj}
\DeclareMathOperator{\map}{map}
\DeclareMathOperator{\clip}{clip}
\newcommand{\clipped}[1]{\underline{\bar #1}}
\newcommand{\Clipped}[1]{\underline{\overline #1}}

\title{SMLE: Safe Machine Learning via Embedded Overapproximation}
\author{
    Matteo Francobaldi\textsuperscript{\rm 1},
    Michele Lombardi\textsuperscript{\rm 1}
}
\affiliations{
    \textsuperscript{\rm 1}DISI, University of Bologna\\

    \{matteo.francobaldi2,michele.lombardi2\}@unibo.it
}

\begin{document}

\maketitle

\begin{abstract}
Despite the extent of recent advances in Machine Learning (ML) and Neural Networks, providing formal guarantees on the behavior of these systems is still an open problem, and a crucial requirement for their adoption in regulated or safety-critical scenarios.
We consider the task of training differentiable ML models guaranteed to satisfy designer-chosen properties, stated as input-output implications.
This is very challenging, due to the computational complexity of rigorously verifying and enforcing compliance in deep neural models.
We provide an innovative approach based on three components: 1) a general, simple architecture enabling efficient verification with a conservative semantic; 2) a rigorous training algorithm based on the Projected Gradient Method; 3) a formulation of the problem of searching for strong counterexamples.
The proposed framework, being only marginally affected by model complexity, scales well
to practical applications, and produces models that provide full property satisfaction guarantees.
We evaluate our approach on properties defined by linear inequalities in regression, and on mutually exclusive classes in multilabel classification.
Our approach is competitive with a baseline that includes property enforcement during preprocessing, i.e. on the training data, as well as during postprocessing, i.e. on the model predictions.
Finally, our contributions establish a framework that opens up multiple research directions and potential improvements.
\end{abstract}

%

\section{Introduction}
\label{sec:intro}
\input{intro}

\section{Related Work}
\label{sec:related}
\input{related}

\section{Robust Training Framework}
\label{sec:formalization}
\input{formalization}

\section{Framework Groundings}
\label{sec:groundings}
\input{groundings}

\section{Experimentation}
\label{sec:experimentation}
\input{experimentation}

\section{Conclusion}
\label{sec:conclusion}
\input{conclusion}

\bibliography{aaai25}
\newpage
\input{supplemental_arxiv}
\end{document}

%% file: intro.tex
Recent years have seen a rapid expansion in the deployment of AI and Machine Learning (ML) systems, so that their robustness and safety have become a matter of public concern.
In safety-critical or regulated contexts such as automation, healthcare, and risk assessment, AI solutions must comply with specific properties set by designers.
In non-critical settings, the ability of AI systems to meet user expectations is still an important factor for their acceptance.
The AI act recently passed by the European Union is considered by many as the first of many legal frameworks that will stress the importance of compliance for AI systems in high-risk sectors.

However, training robust models is challenging, for multiple reasons.
The theoretical basis for most training formulations
allows for some level of error and uncertainty.
It's almost impossible to ensure that the training samples are fully representative of real-world usage, leading to unpredictable behavior when out-of-distribution inputs are encountered.
Finally, a body of research \cite{chakraborty2021survey,tseng2024can} suggests that complex AI models can be fragile and susceptible to adversarial attacks.

Several research directions addressing these issues have emerged, but the problem remains largely open.
Verification approaches focus on checking compliance for a given property \cite{liu2021algorithms}, but for most architectures this problem is NP-hard and very difficult.
Other techniques introduce loss terms linked to undesirable properties (e.g. discrimination), or by adversarial training, which generates and accounts for counterexamples at training time \cite{muhammad2022survey}; both methods struggle to provide out-of-distribution guarantees, either for structural reasons or due to the computational complexity of generating and resolving counterexamples.
Some methods enforce property compliance  on the training data;
others work at inference time, e.g. via input randomization \cite{cohen2019certified} or by adjusting the model output to ensure property satisfaction \cite{yu2022towards}.

We tackle the problem of training ML models that are robust w.r.t. formal properties specified as implications.
To this end, we introduce \emph{a general neural architecture} that is simple to implement and enables efficient verification with a conservative semantic.
The architecture is referred to as Safe ML via Embedded overapproximation (SMLE), since it works by augmenting a backbone network with a low-complexity, trainable overapproximator.
For certain classes of properties and design choices, conservative verification of SMLE networks has polynomial time complexity.

We use SMLE as a building block for \emph{a training framework based on the Projected Gradient Method}.
Our method \emph{guarantees satisfaction of the desired properties upon convergence}.
While the computational cost is noticeable, it is \emph{only marginally affected by the backbone network} and remains affordable for practical-scale problems.
We ground our framework for two classes of properties.
First, we consider properties defined via \emph{linear inequalities}, a large class with applications such as risk assessment, stable time-series forecasting, collision avoidance.
Second, we consider a much more complex (but more specific) combinatorial property, namely \emph{mutual exclusions in multi-label classification}.
We evaluate our approach on synthetic and real-world datasets, considering both random and realistic properties.
As a strong baseline for the comparison, we use pre-processing and post-processing solutions relying on exact maximum-a-posteriori computation.
While our method shows slightly decreased accuracy compared to the baselines, it consistently achieves safety guarantees without increasing the complexity of the inference process.
Finally, our contributions open up multiple research directions.

%% file: related.tex
Two trends mainly arise from the literature on AI safety and robustness: verification and robust model training. 

Verification methods attempt to formally certify the validity of properties in already-trained models. They rely either on Optimization and Searching or on Reachability Analysis.
The former work in a declarative fashion: they encode the ML system into a chosen modeling language, such as  Mixed-Integer Linear Programming \cite{tjeng2017evaluating, bunel2018aunified, fischetti2018deep, anderson2020strong, tsay2021partition} or Satisfability Modulo Theory \cite{ehlers2017formal, huang2017safety, katz2017reluplex, katz2019marabou}, hence they search for a counterexample that falsifies the assertion. These methods can perform exact verification, but usually fail to scale to real-world use cases.
The latter adopt instead algorithmic approaches: they analyze the layer-by-layer propagation of a set of inputs through a neural network, in order to reconstruct the set of reachable outputs, hence to check whether any of them falls into an unsafe region. By maintaining an outer-approximation of the input set during the propagation \cite{singh2018fast, xiang2018output, gehr2018ai2, li2019analyzing, singh2019abstract}, these methods offer increased tractability, but at the price of a loss of completeness. A comprehensive survey on AI Verification is provided in \cite{liu2021algorithms}. 

Robust training methods, on the other hand, address safety and robustness in ML systems directly in the design of the model. Some of these methods work in a post-processing fashion, by equipping the main model with auxiliary mechanisms that operate right before or immediately after the actual inference. The former detect and purify, or more conservatively reject, malicious inputs \cite{dhillon2018stochastic, samangouei2018defensegan, yang2019menet, pang2018towards, metzen2017detecting, xu2017feature}; the latter correct the output to enforce constraint satisfaction \cite{WABERSICH2021109597,
yu2022towards}. Another class of methodologies, widely recognized as one of the most effective to improve (local) robustness, is the so-called Adversarial Training \cite{madry2018towards, zhang2019you, shafahi2019adversarial, wang2020Improving, zhang2020attacks, wong2020Fast, kim2021understanding}. Here the model is not trained over the original, clean datapoints, but over their worst adversarial examples, or over a combination of the two. The main challenge of adversarial training arises from the generation of adversarial examples, an NP-Hard problem that should be solved iteratively during the training loop. Thus, the focus of this research branch is on designing clever ways to approximate this problem to speed up the computations. A clear overview on Adversarial Training is proposed in \cite{ijcai2021p591}.

The key difference between verification and robust model training is that verification offers formal guarantees when a property is satisfied but provides no guidance for correcting the model when it is violated, while robust training actively promotes property satisfaction but lacks formal certification. This work aims to bridge the gap between these two lines, by introducing a methodology to enforce and formally guarantee the satisfaction of properties at training time.

%% file: formalization.tex
This section introduces our framework for training differentiable ML models with satisfaction guarantees, starting with a formalization and analysis of the target problem.

\paragraph{Formal Properties and Robust Training}

Let $X$ and $Y$ be random variables respectively representing the model input and the quantity to be predicted; let $\mathcal{X}, \mathcal{Y}$ be their supports -- assumed to be bounded -- and $P(X, Y)$ their joint distribution.
Finally, let $f(x; \theta)$ be a deterministic and differentiable ML model, such as a neural network, with parameter vector $\theta$.
We consider properties stated as implications in the form:
\begin{align}
    \label{eq:property}
    \forall x \in \mathcal{X}, Q(x) \Rightarrow R(f(x; \theta))
\end{align}
where $Q$ and $R$ are logical predicates defined respectively over $\mathcal{X}$ and $\mathcal{Y}$. 
This class of properties includes consistency in classification (e.g. ``a dog is also an animal'', mutual exclusive or forbidden labels), bounding the variability of predictions in multi-step time series forecasting, and safety properties for collision avoidance systems.
More examples can be found in the VNN competition \cite{brix2023fourth}.

We will not consider properties defined via local perturbations, such as classical adversarial examples or local monotonicity.
While predicates modeling these properties for a \emph{fixed} set of examples can be constructed, true robustness in these settings should account for the actual input distribution $P(X)$, which is typically inaccessible.
While this limitation is common to all robustness methods, it is especially at odds with our approach, which emphasizes full guarantees.


Training a robust model, then, amounts to solving:
\begin{align}
    \label{eq:training}
    \argmin_\theta \left\{
    \mathrm{E}_{x, y \sim P(X, Y)} \left[L(y, f(x; \theta) \right]
    \text{ s.t.  \cref{eq:property}}
    \right\}
\end{align}
where $L$ is the loss function for an individual example and the expectation is usually approximated via the sample average over the training set.
\Cref{eq:training} is a constrained optimization problem with a differentiable loss; it can be solved to local optimality, for example, via the Projected Gradient Method \cite{parikh2014proximal}.
This approach pairs every gradient update with a geometrical projection in feasible space.
Formally, the model parameters are updated via:
\begin{align}
    \theta^{(k+1)} = \proj_{f} (\theta^{(k)} - \eta^{(k)} \cdot \nabla \tilde{L}(\theta^{(k)}) )
\end{align}
where the superscripts refer to $k$-th iteration, $\tilde{L}$ is the expectation from \cref{eq:training}, and $\eta^{(k)}$ is the learning rate vector.
The projection operator is defined as:
\begin{align}
    \label{eq:projection}
    \proj_{f} (\theta) = \argmin_{\theta'} \{ \|\theta' - \theta \|_2^2 \text{ s.t. \cref{eq:property} for $\theta'$}\}
\end{align}
Intuitively, we seek the smallest parameter adjustment that guarantees the satisfaction of the desired property.
The main challenge with this approach is that solving \cref{eq:projection} can be very expensive from a computational perspective, due to two main factors: 1) the universal quantification in \cref{eq:property}; 2) the high non-linearity and large size of modern neural models.
In fact, even just checking the validity of \cref{eq:property} is a very difficult NP-hard problem \cite{katz2017reluplex}.

\paragraph{SMLE Architecture}

We address the mentioned difficulties by introducing a general neural architecture that is simple to implement and enables efficient \emph{conservative} verification for \cref{eq:property}.
We start by viewing the ML model $f$ as a composition of an arbitrary embedding function $h$ and a linear (more precisely, affine) output function $g$:
\begin{align}
    f(x; \theta) \equiv g(z; \theta_g) \circ h(x; \theta_h)
\end{align}
Where $z$ is the output of the embedding function and $\theta \equiv (\theta_g, \theta_h)$.
This decomposition is both natural and common for many neural networks -- including classifiers, by focusing on their logit output.
Then, we augment the model by applying clipping to the output of $h$.
In detail, we introduce:
\begin{align}
    \clip(z; l; u) = \max(l, \min(u, z))
\end{align}
which is employed to obtain:
\begin{align}
    \label{eq:smle}
    g(\clipped{z}; \theta_g)
    \circ \clip(z; \underline{h}(x; \theta_{\underline{h}}); \bar{h}(x; \theta_{\bar{h}}))
    \circ h(x; \theta_h)
\end{align}
The embedding $z$ is processed to obtain a clipped embedding $\clipped{z}$, with lower and upper bounds computed by two \emph{auxiliary  models} $\underline{h}$ and $\bar{h}$.
The clipped embedding is then transformed by $g$ to provide the model output.
The auxiliary models can be freely chosen, as long as they are differentiable and significantly simpler than the embedding function $h$.
In the simplest case, $\underline{h}$ and $\bar{h}$ can be constant vectors, but they could be implemented as fully-fledged neural networks.
The structure of our architecture is depicted in \cref{fig:smle_arch}.

\input{images/smle}

The structure of \cref{eq:smle} ensures that the input to the $g$ function is contained in the box:
\begin{equation}
    \Clipped{H}(x; \theta_{\underline{h}}, \theta_{\bar{h}}) = [\underline{h}_1(x), \bar{h}_1(x)] \times \ldots \times [\underline{h}_n(x), \bar{h}_n(x)]
\end{equation}
where $n$ is the size of the embedding vector.
In other words, our models \emph{include a trainable overapproximation}.
For this reason, we refer to the architecture from \cref{eq:smle} as Safe Machine Learning via Embedded overapproximation (SMLE).
Many neural architectures employed in AI research and real-world applications can be easily adapted to this structure.


The SMLE architecture allows for efficient, albeit conservative, property verification.
This can be framed as searching for a counterexample, i.e. an input value that violates \cref{eq:property}:
\begin{align}
    \label{eq:counter}
    \exists x \in \mathcal{X} : Q(x) \wedge \neg R(f(x; \theta))
\end{align}
With the SMLE architecture, the test can be replaced by:
\begin{equation}
    \label{eq:counter_smle}
    \exists x \in \mathcal{X}, \clipped{z} \in \Clipped{H}(x; \theta_{\underline{h}}, \theta_{\bar{h}}) : Q(x) \wedge \neg R(g(\clipped{z}; \theta_g))
\end{equation}
Since $\Clipped{H}$ is by construction an overapproximation, if no counterexample for \cref{eq:counter_smle} can be found, then the SMLE model is guaranteed to satisfy the desired property.
If a counterexample is found, however, the test is inconclusive, so that the procedure is incomplete and has a \emph{conservative} semantic.

The main appeal of \cref{eq:counter_smle} is that is involves only components whose complexity can be \emph{freely controlled}.
Regardless of the complexity of the embedding function $h$, by choosing sufficiently simple $\underline{h}$, $\overline{h}$, and $g$, conservative verification can be made efficient \emph{by construction}.
For example, if $\underline{h}, \bar{h}$ are linear, and $Q$ and $R$ represent polytope membership tests, then \cref{eq:counter_smle} is a linear system of inequalities that can be solved in polynomial time.
The main drawback is the loss of completeness, which is however less problematic when the focus is on enforcing property satisfaction at training time.

\paragraph{Conservative Projection}

The SMLE architecture provides the basis for building an efficient, conservative, version of the projection operator from \cref{eq:projection}.
As a first step in this direction, we restrict projection to the output parameters $\theta_g$ and replace \cref{eq:property} with its conservative SMLE version:
\begin{subequations}
\label{eq:cons_proj}
\begin{align}
    & \argmin_{\theta_g'} \|\theta_g' - \theta_g\|_2^2 \\
    & \quad \text{s.t. } R(g(\clipped{z}; \theta_g')) \qquad \forall x \in \mathcal{X} : Q(x), \forall \clipped{z} \in \Clipped{H}(x)
    \label{eq:cons_proj2}
\end{align}
\end{subequations}
In the formulation, the parameters $\theta_{\underline{h}}, \theta_{\bar{h}}$ are omitted from $\Clipped{H}$ since they are not affected by projection.
\Cref{eq:cons_proj} is considerably simpler to solve than the original projection operator, since it involves only simple functions and much fewer parameters.
The problem remains challenging, however, since it contains an infinite number of constraints due to the use of universal quantification.

\input{algos/delayed_constraint_projection}

We address this issue via a delayed constraint generation approach.
We iteratively introduce constraints associated to counterexamples, which can be efficiently found thanks to the SMLE architecture.
Given a finite collection of counterexamples $C$, \cref{eq:cons_proj} is then replaced by:
\begin{equation}
    \label{eq:cons_proj_finite}
    \argmin_{\theta_g'} \{ \|\theta_g' - \theta_g\|_2^2 \text{ s.t. }  R(g(\clipped{z}; \theta_g')), \forall \clipped{z} \in C \}
\end{equation}
For improved efficiency, and without loss of generality, we restrict our attention to the value of the clipped embedding $\clipped{z}$ of each counterexample $(x, \clipped{z})$, since its input $x$ does not appear in the body of \cref{eq:cons_proj2}.

This Projection with Delayed Constraint Generation process is outlined in \cref{alg:pdcg}.
As the convergence properties of the process have not yet been analyzed in detail, we introduce two mitigation strategies to ensure that the algorithm cost is manageable.
First, to reduce the memory requirements, we store a finite number of counterexamples, discarding older ones according to a FIFO policy.
Second, we enforce an iteration limit: if this is reached, the property might not be satisfied and the projection is deemed not entirely successful, though the model weights are still updated.

\paragraph{Counterexample Generation}

In principle, counterexample generation in \cref{alg:pdcg} could be efficiently handled by solving \cref{eq:counter_smle}.
In practice, however, that leaves no control on which kind of counterexample is obtained: if these are too weak, the projection process could become unreasonably slow.
Unfortunately, defining ``strong'' counterexamples is non-trivial, especially without making assumptions on the nature of the property to be satisfied.
In particular, since we treat $Q$ and $R$ as logical predicates, they have no associated continuous measure of constraint violation.

We argue that stronger counterexamples are those requiring large parameter adjustments for their resolution.
Hence, given a pair $(x, \clipped{z})$ satisfying $Q(x)$, we propose to define its strength as a counterexample by means of the L1 norm:
\begin{equation}
    \label{eq:ce_strength}
    \|\theta_g^* - \theta_g \|_1
\end{equation}
where $\theta_g$ is the current weight vector for the $g$ function and $\theta_g^*$ is defined as:
\begin{equation}
    \label{eq:ce_fix}
    \theta_g^* = \argmin_{\theta_g'}
    \left\{ \|\theta_g' - \theta_g\|_2^2
    \text{ s.t. }
    R(g(\clipped{z}; \theta_g'))
    \right\}
\end{equation}
For a pair $x, \clipped{z}$ that already satisfies the property, we have $\theta_g^* - \theta_g = 0$, while we have $\|\theta_g^* - \theta_g\|_1 > 0$ for a true counterexample.
In \cref{eq:ce_strength} we use the L1, rather than L2, norm for its lower computational complexity.
To the same end, we also propose to restrict the type of adjustment used to evaluate the counterexample strength.
In particular, since $g$ is an affine function, by restricting the allowed changes to those affecting the offset (i.e. translation), we have that:
\begin{align}
    g(\clipped{z}; \theta_g') =
    g(\clipped{z}; \theta_g + \delta) =
    g(\clipped{z}; \theta_g) + \delta 
\end{align}
where $\delta$ is the difference $\theta_{g}' - \theta_{g}$.
We now frame the problem of generating a strong counterexample as that of finding \emph{a pair $x, \clipped{z}$ whose resolution requires the largest translation}:
\begin{subequations}
\label{eq:smle_counter_max}
\begin{align}
    x^*, \clipped{z}^* & = \argmax_{x, \clipped{z}} \|\delta^*\|_1\\
    \text{s.t.: }
    & \delta^* = \argmin_{\delta} \{ \|\delta\|_2^2 \text{ s.t. }
    R(g(\clipped{z}; \theta_g) + \delta) \} \label{eq:smle_counter_fix}\\
    & 
    Q(x) \wedge \clipped{z} \in \Clipped{H}(x)
\end{align}
\end{subequations}
where $\|\delta^*\|_1$ and $\|\delta'\|_2^2$ are equivalent to $\|\theta_g^* - \theta_g \|_1$ and $|\theta_g' - \theta_g \|_2^2$, since we are restricted to translation.
While translation alone might be insufficient to solve the projection from \cref{eq:projection}, here we are interested in resolving \emph{a single} pair $x, \clipped{z}$. This can always be done by translation if the $R$ predicate has at least one positive assignment.
Moreover, we have that $\|\delta\|_1 = 0$ iff no counterexample exists.

The formulation from \cref{eq:smle_counter_max} is partially heuristic in nature, therefore further simplifications can be done for specific properties if they bring computational or quality advantages.
In any case, the approach is fairly general, built on a solid rationale, and should result in strong counterexamples.
As a challenge, the equation describes a bilevel optimization problem for which defining a solution approach is non-trivial.
We will discuss specific groundings of this formulation for two practically relevant settings in the next section.


\paragraph{Robust Training Algorithm}

We can now introduce our robust training procedure, described in \cref{alg:rt}.
We start by training a SMLE architecture via Stochastic Gradient Descent (SGD) as usual in deep learning, but after every gradient update we perform a single iteration of our projection algorithm.
This phase is meant to improve the model accuracy, while accounting for the need to satisfy the desired property without an excessive computational cost.
Once convergence is reached according to usual SGD criteria, we attempt a full projection.
If the process succeeds, the resulting SMLE model is guaranteed to satisfy the property, without any further intervention.
One goal of our empirical evaluation is investigating the success rate or \cref{alg:rt}.

\input{algos/robust_training}

%% file: images/smle.tex
\begin{figure}[bt]
    \centering
    \includegraphics[width=0.9\linewidth]{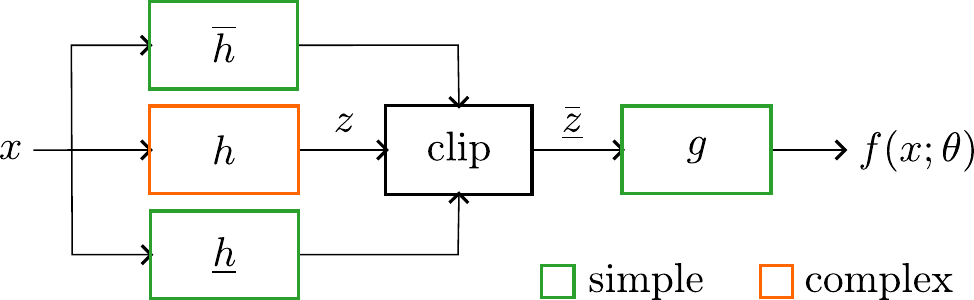}
    \caption{A depiction of a SMLE architecture.}
    \label{fig:smle_arch}
\end{figure}

%% file: algos/delayed_constraint_projection.tex
\begin{algorithm}[tb]
\caption{\textsc{pdcg}$(\theta_g, n_{it}, n_{xs})$}
\label{alg:pdcg}
\begin{algorithmic}
    \STATE $C = $ empty queue
    \FOR{$k=1..n_{it}$}
        \STATE search for a counterexample $(x^*, \clipped{z}^*)$
        \IF{no counterexample is found}
            \RETURN success
        \ENDIF
        \IF{$|C| = n_{xs}$}
            \STATE remove the first example in $C$
        \ENDIF
        \STATE append $\clipped{z}^*$ to $C$
        \STATE obtain $\theta_g^{(k)}$ by solving \cref{eq:cons_proj_finite}
    \ENDFOR
    \RETURN failure
\end{algorithmic}
\end{algorithm}

%% file: algos/robust_training.tex
\begin{algorithm}[tb]
\caption{\textsc{rt}$(\text{\it training-data}, \text{\it SGD-params}, \theta, n_{it}, n_{xs})$}
\label{alg:rt}
\begin{algorithmic}
    \FOR{every usual SGD iteration}
    \STATE perform a gradient descent update
    \STATE run $\text{\sc pdcg}(\theta^{(k)}, 1, n_{xs})$
    \ENDFOR
    \RETURN $\text{\sc pdcg}(\theta^{(k)}, n_{it}, n_{xs})$
\end{algorithmic}
\end{algorithm}

%% file: groundings.tex
Grounding our framework for a given setting requires:
1) choosing a SMLE architecture;
2) defining an implementable formulation for the projection problem from \cref{eq:cons_proj_finite}; and 3) doing the same for the counterexample generation problem from \cref{eq:smle_counter_max}.
While the first step is generally easy, the latter two are non-trivial and depend on the considered properties.
Here, we discuss viable choices for two settings: 1) the case where $Q$ and $R$ are defined via linear inequalities; and 2) mutual exclusions in multi-label classification.
We will focus on mathematical formulations, but obtaining a formally correct grounding requires also to account for solver tolerances and floating-point error propagation.
Tolerance-aware formulations can be found in the supplemental material, but we leave robust handling of floating point errors for future work.

\paragraph{Linear Inequalities}

This setup was chosen since it captures a wide range of practically relevant properties \cite{brix2023fourth}.
In this case, $Q$ and $R$ represent polytope membership tests and are defined via the inequalities:
\begin{subequations}
\label{eq:reg_property}
\begin{align}
    Q(x) & \equiv Q x \leq q \label{eq:reg_property_in}  \\
    R(\hat{y}) & \equiv R \hat{y} \leq r \label{eq:reg_property_out}
\end{align}
\end{subequations}
Where $\hat{y}$ is the model output, $Q, R$ are coefficient matrices, and $q, r$ are the corresponding right-hand side vectors.
The projection problem from \cref{eq:cons_proj_finite} corresponds to:
\begin{subequations}
\label{eq:polytopes_projection}
\begin{align}
    \argmin_{\theta_g'} \
    & \|\theta_g' - \theta_g\|_2^2 \\
    \text{s.t. }
    & R \hat{y}_i \leq r & \forall \clipped{z}_i \in C \\
    & \hat{y}_i = \theta_{g,0} + \theta_{g,1:n} \, \clipped{z}_i & \forall \clipped{z}_i \in C \label{eq:haty_def}
\end{align}
\end{subequations}
where $\theta_{g,0}$ is the offset vector and $\theta_{g,1:n}$ the coefficient matrix for the affine transformation $g$.
\Cref{eq:polytopes_projection} is a poly-time solvable Quadratic Program.

The process for obtaining a formulation for the counterexample generation problem is more involved and can be found in the supplemental material.
The key observation is that resolving a counter example $(x, \clipped{z})$ violating a given linear inequality requires a translation that is proportional to the classical Linear Programming notion of constraint violation.
Accordingly, we generate counterexamples by solving:
\begin{subequations}
\label{eq:polytopes_counter}
\begin{align}
    \argmax_{x, \clipped{z}} \ & \sum_{k=1}^K \max(0, R_k \hat{y} - r_k) \label{eq:polytopes_counter_obj} \\
    \text{ s.t. }
    & Q(x) \\
    &\clipped{z} \geq \underline{h}(x; \theta_{\underline{h}}) \\ 
    &\clipped{z} \leq \max(\underline{h}(x; \theta_{\underline{h}}), \bar{h}(x; \theta_{\bar{h}})) \label{eq:polytopes_counter_const} \\
    & \hat{y} = \theta_{g,0} + \theta_{g,1:n} \, \clipped{z}
\end{align}
\end{subequations}
where $K$ is the number of rows in $R$.
The problem can be stated as Mixed Integer Linear Program by linearizing the $\max$ operator, which in \Cref{eq:polytopes_counter_obj} ensures that non-violated constraints are correctly treated as not needing resolution, while in \Cref{eq:polytopes_counter_const} it models the degeneracy of the clipping operation, occurring when $\underline{h}(x; \theta_{\underline{h}}) \geq \bar{h}(x; \theta_{\bar{h}})$.
Due to simplifications made in our grounding process, \cref{eq:polytopes_counter} is not a equivalent to \cref{eq:smle_counter_max}.
However, the problem is still capable of generating strong counterexamples, and the objective is still 0 if no counterexample exists.

\paragraph{Mutual Exclusive Classes}
The second setup find applications when tagging content, or when determining the traits of biological samples, if certain tag or trait combinations should never be predicted together. 
It was chosen as an example of a \emph{combinatorial} property in a classification setting, which comes with unique challenges.
Since SMLE assumes the last layer of the architecture is linear, the $Q$ and $R$ predicates are defined in this case as:
\begin{subequations}
\label{eq:class_property}
\begin{align}
    Q(x) & \equiv \text{\sc true} \label{eq:class_property_in} \\
    R(\hat{y}) & \equiv I^+(\hat{y}_h) + I^+(\hat{y}_k) \leq 1 & \forall h, k \in F \label{eq:class_property_out} 
\end{align}
\end{subequations}
where $\hat{y}$ represents the logit output of the multilabel classifier, $F$ is the set of mutually exclusive class pairs, and the indicator funtion $I^+(\hat{y}_h) = \text{\sc true}$ iff $\hat{y}_h \geq 0$ and $\text{\sc false}$ otherwise.
In fact, if a sigmoid layer is used to obtain the actual classifier output, class $h$ will predicted as true iff $\hat{y}_h \geq 0$.
The projection problem from \cref{eq:cons_proj_finite} corresponds to:
\begin{subequations}
\label{eq:mutex_projection}
\begin{align}
    & \argmin_{\theta_g'}
    \|\theta_g' - \theta_g\|_2^2 \\
    &\quad \text{s.t. }
    I^+_{i,h} + I^+_{i,k} \leq 1 & \forall h, k \in F, \forall \clipped{z}_i \in C \\
    &\quad M I^+_{i,k} \geq \hat{y}_{i,k} & \forall k \in O, \forall \clipped{z}_i \in C \\
    &\quad \hat{y}_i = \theta_{g,0} + \theta_{g,1:n} \, \clipped{z}_i & \forall \clipped{z}_i \in C \\
    &\quad I^+_{i,k} \in \{0, 1\} & \forall k \in O, \forall \clipped{z}_i \in C
\end{align}
\end{subequations}
where $O$ is the set of output classes and auxiliary variables are used to model the $I^+(\hat{y}_{i,k})$ predicates; $M$ is a sufficiently large constant that is used to ensure that $\hat{y}_{i,k} < 0$ if $I^+_k = 0$.
Ways to choose the value of $M$ and to handle strict inequalities are discussed in the supplemental material.
\Cref{eq:mutex_projection} is a Mixed Integer Quadratic Programming problem, which modern mathematical programming solvers are capable of addressing.

We propose counterexample generation problem based on the observation that, if mutually exclusive classes $h$ and $k$ are predicted at the same time, then adjusting any of $\hat{y}_h$ or $\hat{y}_k$ so that it is less than 0 resolves the violation.
\begin{subequations}
\label{eq:mutex_counter}
\begin{align}
    &\argmax_{x, \clipped{z}} 
    I^{m}_{h,k} \min(\hat{y}_h, \hat{y}_k) \\
    &\quad \text{s.t. }
    Q(x) \\
    &\quad \clipped{z} \geq \underline{h}(x; \theta_{\underline{h}}) \\ 
    &\quad \clipped{z} \leq \max(\underline{h}(x; \theta_{\underline{h}}), \bar{h}(x; \theta_{\bar{h}})) \\
    &\quad \hat{y} = \theta_{g,0} + \theta_{g,1:n} \, \clipped{z} \\
    &\quad MI^+_{k} - M \leq \hat{y}_{k} & \forall k \in O \label{eq:class_def_counter} \\
    &\quad I^m_{h,k} \leq \frac{1}{2}(I^+_{h} + I^+_{k}) & \forall h, k \in F \label{eq:mutex_indicator}\\
    &\quad I^+_{k} \in \{0, 1\} & \forall k \in O \\
    &\quad I^m_{h,k} \in \{0, 1\} & \forall h, k \in F
\end{align}
\end{subequations}
\Cref{eq:mutex_indicator} ensures that the additional binary variables $I^m_{h,k}$ can be set to 1 only if mutually exclusive classes $h$ and $k$ are predicted. \Cref{eq:class_def_counter} ensures that the indicator variables $I^+_k$ can be set to 1 only if the corresponding logit otuput is at least 0.
The $\min$ operator in the objective can be linearized by introducing a fresh continuous variable and two inequalities, as shown in the supplemental material.
\Cref{eq:mutex_counter} is Mixed Integer Linear Program and matches the semantic from \cref{eq:smle_counter_max}, since adjusting $\hat{y}_k$ so that it equals 0 corresponds to choosing $\delta_k = - \hat{y}_k$.

%% file: experimentation.tex

We conduct an empirical study designed around the following research questions:
\emph{(Q1, Accuracy)} Can our approach achieve an acceptable prediction accuracy, while providing full guarantees?
\emph{(Q2, Guarantees)} How effective is our method at providing guarantees compared to existing approaches?
\emph{(Q3, Ablation Study)} How is our framework accuracy impacted by its key hyperparameters?

\paragraph{Compared Approaches}
The baseline for our comparison relies on a \emph{maximum-a-posteriori} operator, defined as:
\begin{align}
\label{eq:outputprojection}
\map(y) = \argmax_{y'} \{ \mathcal{L}(y'\mid y) \text{ s.t. } R(y')\}
\end{align}
where $\mathcal{L}$ denotes the likelihood of an adjusted prediction $y'$ w.r.t. a reference prediction $y$.
Intuitively, we correct an infeasible output $y$, by projecting it to its closest feasible point. 

Our baseline consists of two competitors, \texttt{preprocess} and \texttt{postprocess}. 
In \texttt{preprocess} we apply $\map$ to enforce property satisfaction in the training labels, then we train the model on this modified dataset. 
In \texttt{postprocess} we apply $\map$ to enforce the property in the model predictions at inference time.
When using \texttt{preprocess}, the highest computational cost, i.e. property enforcement, is paid offline: then, any learning algorithm can be used to obtain a model with fast inference. As a disadvantage, this approach does not guarantee on the validity of the property. Notably, the resulting model is vulnerable to adversarial attacks.
On the other hand, \texttt{postprocess} provides full satisfaction guarantees, but might be very inefficient at inference time; in the case of mutually exclusive classes, for example, \cref{eq:outputprojection} is an NP-Hard problem that should be solved for each input.

As an optimistic reference, we consider a theoretical approach, \texttt{oracle}, which works by applying $\map$ to the ground truth labels.
We use \texttt{oracle}: 1) to obtain an upper bound on the achievable accuracy; and 2) to quantify the difficulty of satisfying a given property on a dataset, by measuring the accuracy drop caused by the application of $\map$.



Our approach \texttt{smle} is trained through \cref{alg:rt}, after having enforced the property on the training data, as for \texttt{preprocess}.
This pre-training step was not adopted for \texttt{postprocess}, since it did not provide any significant advantage in that case, as revealed by preliminary experiments.

All our experiments, methods and benchmarks are implemented in Python, by relying on the libraries TensorFlow \cite{tensorflow2015-whitepaper}, Keras \cite{chollet2015keras} and Scikit-Learn \cite{scikit-learn} for the ML components, and on the Pyomo \cite{hart2011pyomo, bynum2021pyomo} and OMLT modeling tools \cite{ceccon2022omlt, zhang2024augmenting} and on the Gurobi solver \cite{gurobi}, for the optimization ones.
The implementation details are provided in our code, which is publicly available together with our data\footnote{https://github.com/Francobaldi/SMLE\_AAAI2025}.


\paragraph{Benchmarks}
We cover the two groundings described above, regression and multi-label classification, by adopting two benchmarks for the former, called \emph{synthetic regression} and \emph{multi-step time series forecasting}, and one for the latter, simply called \emph{multi-label classification}.

\textbf{Synthetic Regression}. We design 9 learning tasks, consisting in estimating a vector of the form $F_K(x) = ( ( \sum_{j=1}^{n} x_j )^k )_{k \in K}$.
For each task, we train and test on randomly generated data, and we enforce 6 randomly generated linear properties, defined as in \cref{eq:reg_property}.

\textbf{Multi-Step Time Series Forecasting}. We consider the problem of estimating the $m$ consecutive values $y = (s_{t+1}, \dots, s_{t+m})$ of a time series, by observing the $n$ previous values $x = (s_{t-n+1}, \dots, s_{t})$ of the same series $S = (s_t)_{t=0}^{t=N}$.
We train and test on 25 time series selected from a public repository \cite{makridakis2020m4}, each representing a single learning task and split according to a chronological 80\%-20\% criterion.
For each series $S$, we consider 3 properties defined as:
$\forall x, |y_i - y_{i+1}| \leq \Delta_q, \forall i=1, \dots, m-1$
, for $q = 0.90, 0.95, 1.00$, where $\Delta_q$ denotes the q-quantile of the set of deviations $\{|s_t - s_{t+1}|\}_{t=0}^{t=N}$.
These properties, which can be encoded as in \cref{eq:reg_property}, prevent unreasonably high deviations between two consecutive predictions.

\textbf{Multi-Label Classification}. We train and test on 5 datasets selected from a public repository \cite{ucodataset}, each representing a single learning task and split according to a random 80\%-20\% criterion. For each dataset $D$, we consider 3 properties defined as in \cref{eq:class_property} with $F = \{(c_0, c_1) \in O_D \mid \text{freq}_{D}(c_0, c_1) \leq \Delta_q\}$, where $O_D$ represents the set of possible classes, $\text{freq}_{D}(c_0, c_1)$ the frequency at which the pair $(c_0, c_1)$ occurs among the true labels, while $\Delta_q$ denotes the q-quantile of the set of pair frequencies $\{\text{freq}_{D}(c_0, c_1)\}_{c_0, c_1 \in O_D}$, for $q = 0.0, 0.3, 0.6$. These properties prevent the prediction of unusual combinations.

\paragraph{Q1 Results} 
In \Cref{fig:baseline}, we compare the predictive performance of our framework against its competitors over the three benchmarks. We evaluate the models with a value in the $[0,1]$ interval, corresponding to the $R^2$ in regression, and to the \emph{average class accuracy} in classification.
The reported results are obtained on the test sets, and aggregated across the considered properties, sorted by difficulty.

\input{images/baseline}

\Cref{fig:baseline_synthetic} shows that, as expected, the performance of the considered models decreases with the difficulty of the enforced property. Perhaps counter-intuitively, \texttt{preprocess} appears to outperform \texttt{oracle}, especially as the difficulty of the properties increases.
In fact, this is just a side effect of the inability of \texttt{preprocess} to provide full guarantees:
since \texttt{oracle} can be outperformed only by violating the property, this performance gap simply indicates that \texttt{preprocess} is leading to significant violations.
%
These two trends do not evidently arise in \Cref{fig:baseline_forecasting,fig:baseline_classification}. The reason is that, while properties in the Synthetic benchmark are randomly generated, in the other two they are chosen realistically with respect to the data;
as a result, they tend to be more consistent with the natural data distribution.
%

The \texttt{postptocess} approach can achieve a high predictive performance, and provides satisfaction guarantees.
As already discussed, however, the downside is a more complex and computationally expensive inference process.

Finally, our framework achieves very promising results: in most cases, it performs very closely to its competitors, and in some cases even better (\cref{fig:baseline_classification}).
Remarkably, although \Cref{alg:rt} may fail to reach convergence, all \texttt{smle} models presented in this computational study were able to achieve it, and hence to provide full satisfaction guarantees.
This required, in \cref{alg:pdcg}, a memory size of a single counterexample for linear properties, and only 10 counterexamples for mutually exclusive classes.

\paragraph{Q2 Results}

For a trained SMLE model, reduced accuracy is the only price paid to improve robustness.
This is not the case for our competitors, with \texttt{preprocess} being unreliable in terms of property satisfaction and \texttt{postprocess} having a higher computional cost of inference.
%
On the Synthetic benchmark, the most critical for \texttt{preprocess}, we count the percentage of test examples where the desired property is violated by this model.
On the Classification benchmark, the most critical for \texttt{postprocess}, we compute the slowdown, relative to \texttt{preprocess} (the fastest model), of the inference time per test example, for both \texttt{postprocess} and \texttt{smle}. The results of these experiments are reported in \cref{fig:violation_synthetic,fig:inference_classification}, respectively. 

\input{images/violation}

\input{images/inference}

As shown in \Cref{fig:violation_synthetic}, \texttt{preprocess} exhibits a very high violation rate, exceeding 50\% in the worst cases. In contrast, \texttt{smle} guarantees 0\% violation rate on the same properties and tasks.
While \texttt{postprocess} can achieve the same level of guarantees, \Cref{fig:inference_classification} shows that it requires a significant computational effort at inference time, with a slowdown rising up to $2^{13}$ on the most demanding properties.
In contrast, \texttt{smle} is, regardless of property difficulty, only around twice as slow as \texttt{preprocess} at inference time, evidently due to the effect of the introduced overapproximator architecture.



\paragraph{Q3 Results}

Finally, we investigate the impact in terms of predictive capabilities of two key design choices in our framework, namely, the auxiliary models $\underline{h}$, $\bar{h}$ and the backbone model $h$.
In particular, on the regression benchmarks, we compare two overapproximators with different complexities (constant versus linear), as well as two embedding models with different depth and width (small versus large).
On the Forecasting benchmark, we also compare to two different types of backbone (ReLU versus LSTM). 

\Cref{tab:ablation}, where we display the results over the test sets in terms of $R^2$, aggregated across tasks and properties, shows that our framework is quite robust to different design choices, with only small differences reported between each considered setup.
This suggests that defining the hyperparameters for our framework should not be significantly more difficult than for regular scenarios.


\input{tables/ablation_table}

%% file: images/baseline.tex
\begin{figure}[tb]
    \centering
    \begin{subfigure}[b]{0.42\textwidth}
        \centering
        \includegraphics[width=\textwidth]{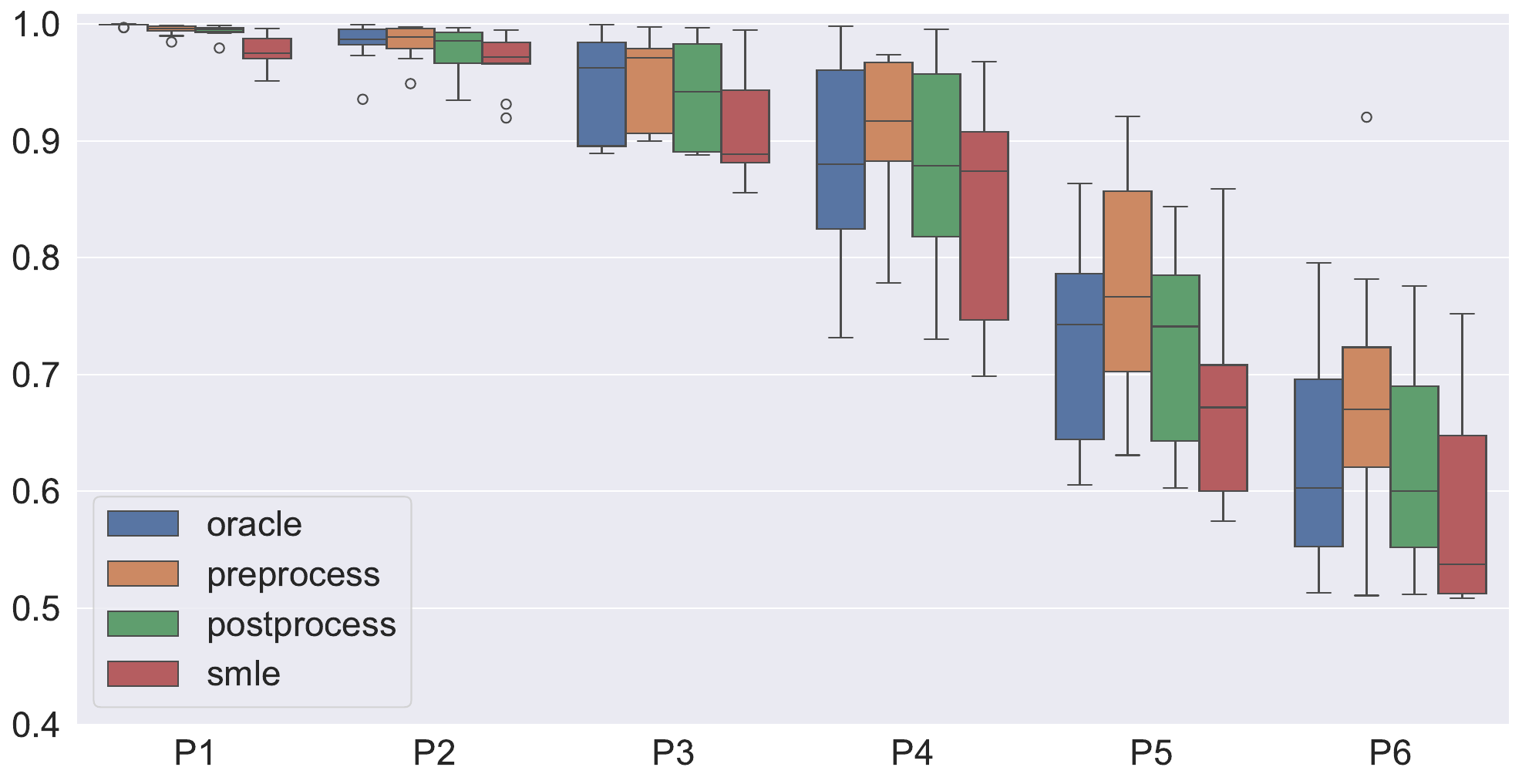}
        \caption{Synthetic}
        \label{fig:baseline_synthetic}
    \end{subfigure}
    
    \begin{subfigure}[b]{0.42\textwidth}
        \centering
        \includegraphics[width=\textwidth]{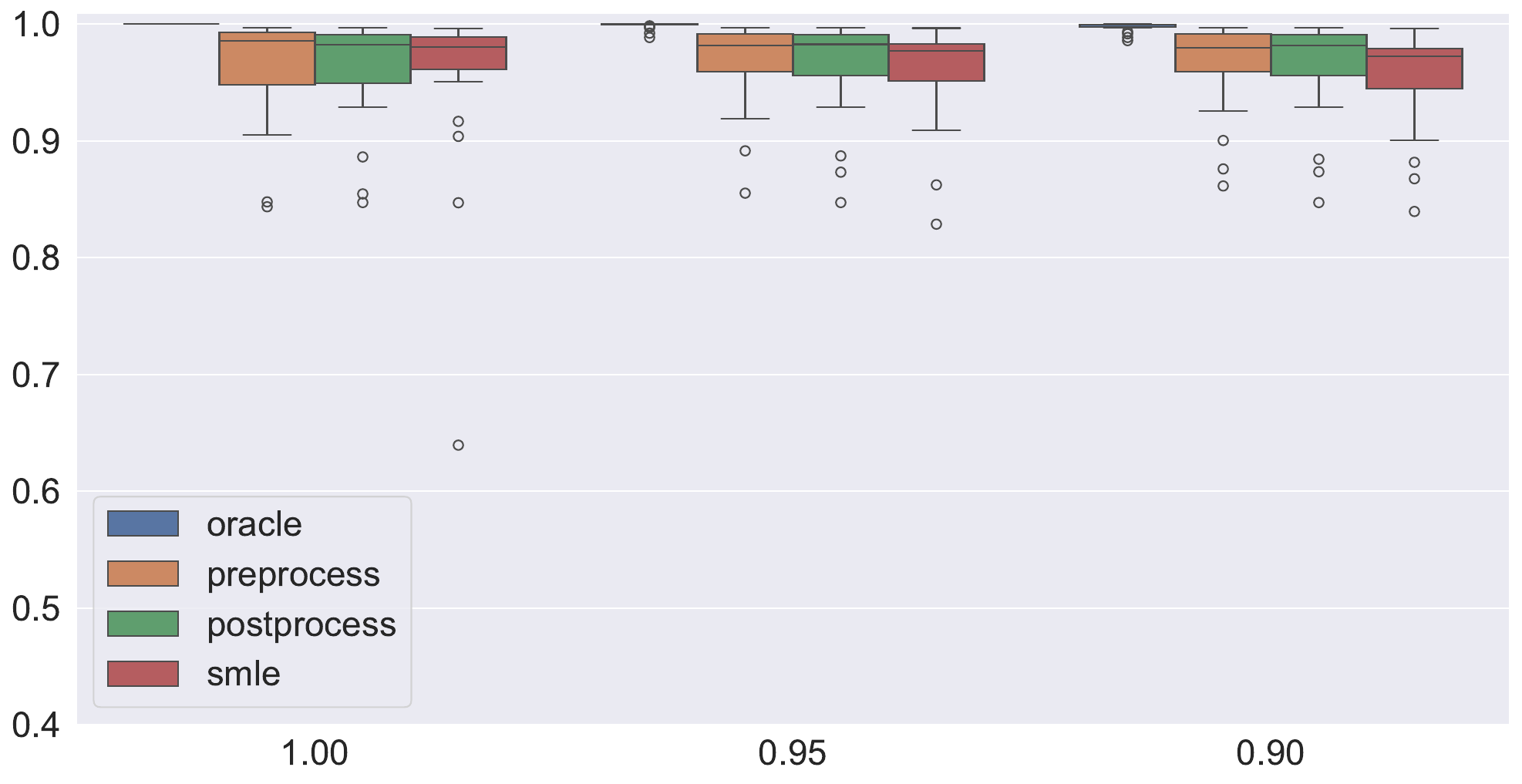}
        \caption{Multi-Step Time Series Forecasting}
        \label{fig:baseline_forecasting}
    \end{subfigure}

    \begin{subfigure}[b]{0.42\textwidth}
        \centering
        \includegraphics[width=\textwidth]{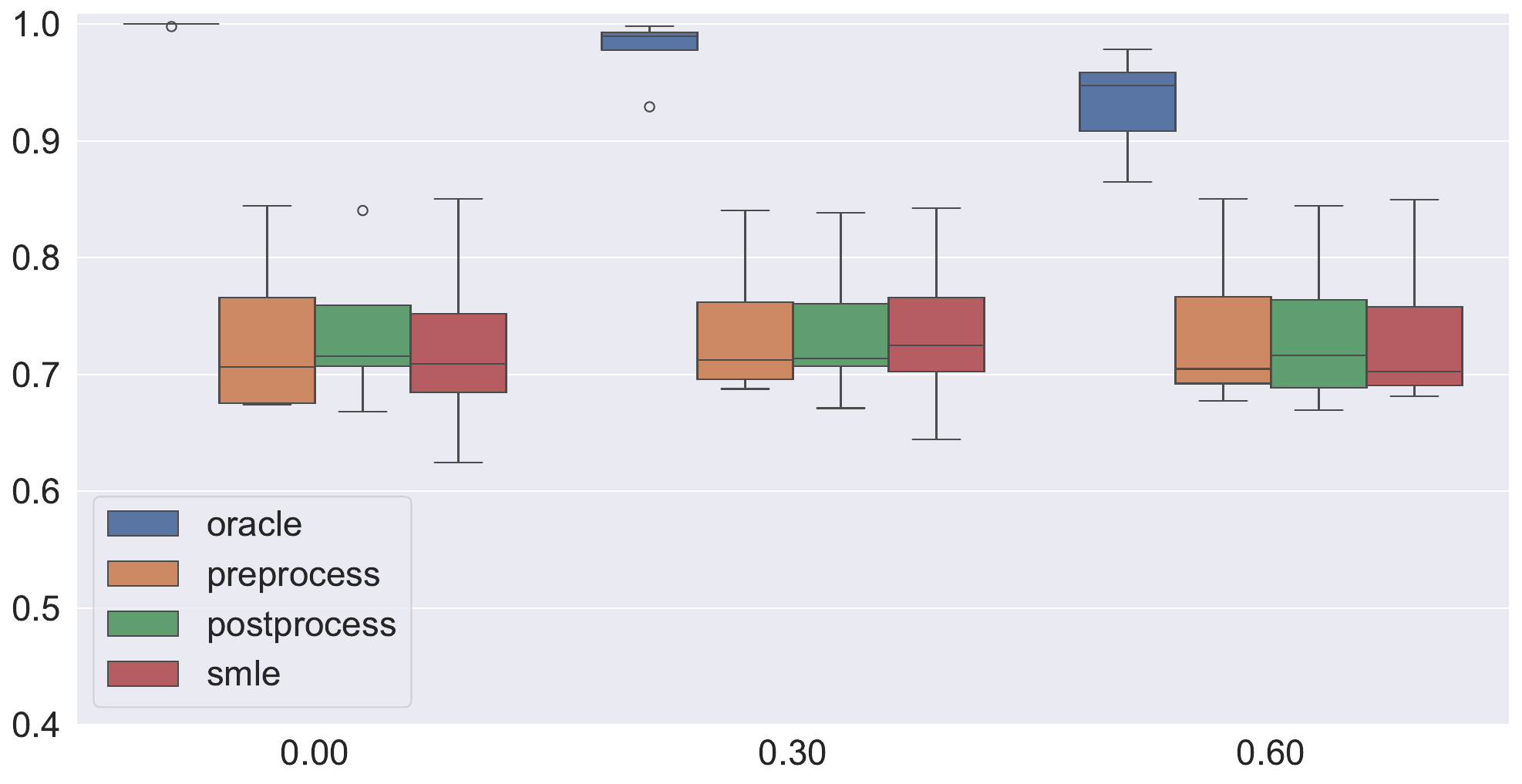}
        \caption{Multi-Label Classification}
        \label{fig:baseline_classification}
    \end{subfigure}
    
    \caption{Baseline results}
    \label{fig:baseline}
\end{figure}

%% file: images/violation.tex
\begin{figure}[tb]
    \centering
    \includegraphics[width=0.42\textwidth]{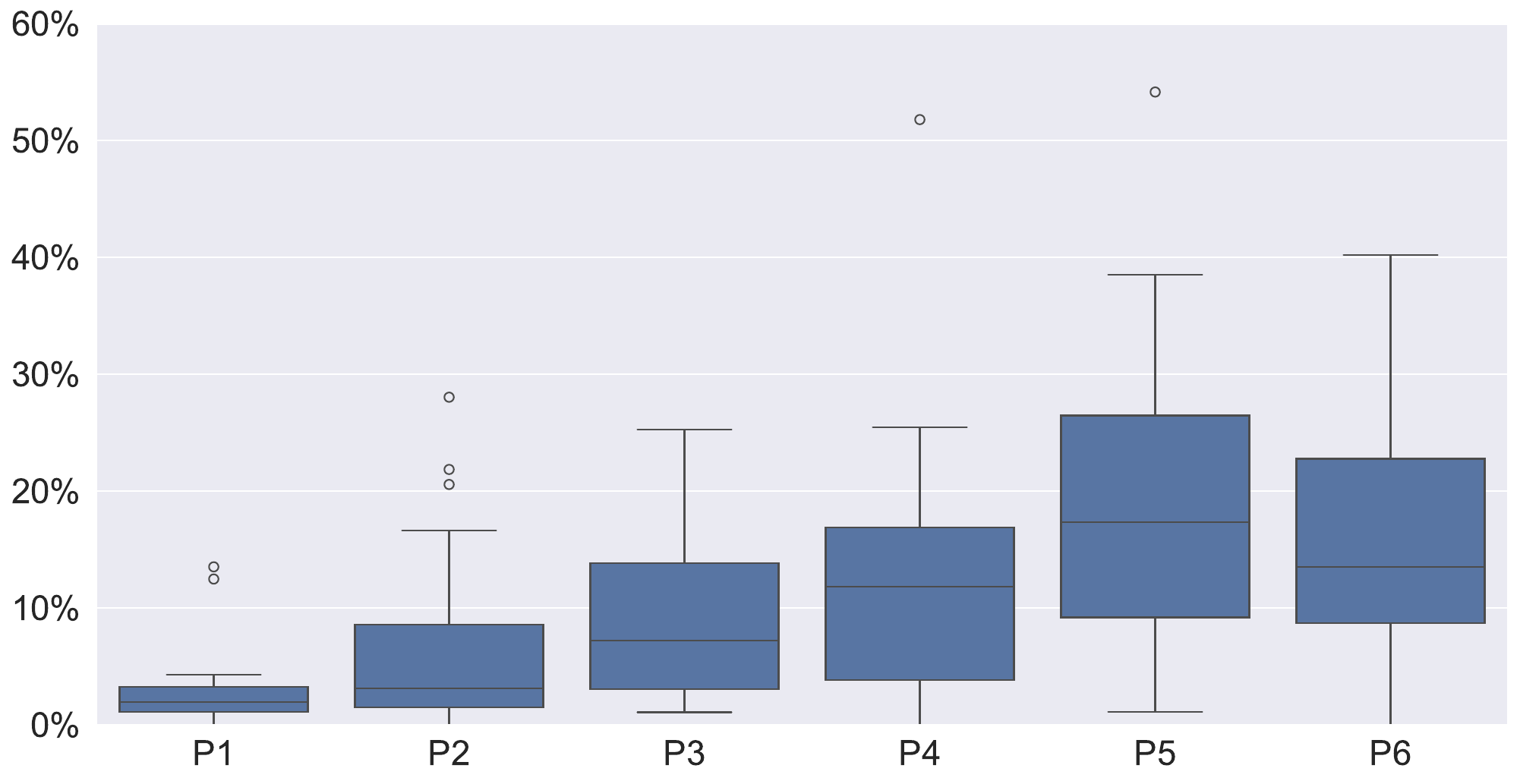}
    \caption{\texttt{preprocess} violation on the Synthetic case}
    \label{fig:violation_synthetic}
\end{figure}

%% file: images/inference.tex
\begin{figure}[tb]
    \centering
    \includegraphics[width=0.42\textwidth]{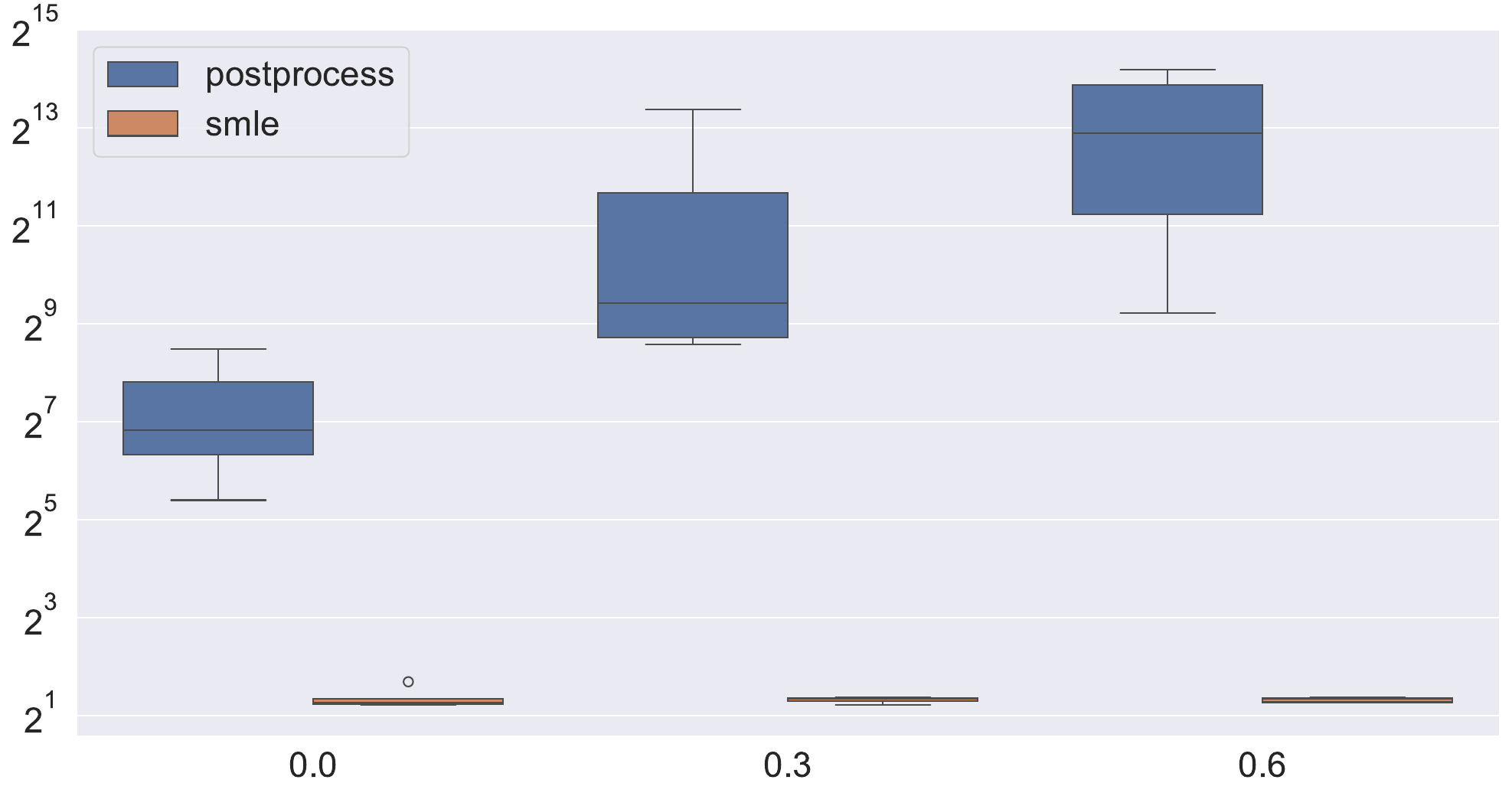}
    \caption{inference slowdown relative to \texttt{preprocess} on the Classification benchmark}
    \label{fig:inference_classification}
\end{figure}

%% file: tables/ablation_table.tex
\begin{table}[tbh]
    \centering
    \renewcommand{\arraystretch}{1.}

    \begin{tabular}{cc|cc}
        \multicolumn{4}{c}{\textbf{Synthetic}} \\ \hline
        \multicolumn{2}{c|}{Aux. Complexity} & \multicolumn{2}{c}{Emb. Size} \\ \hline
        Constant & Linear & Small & Large \\ \hline
        0.814 & 0.819 & 0.819 & 0.823 \\ \hline
    \end{tabular}

    \vspace{0.1cm}
    
    \begin{tabular}{cc|cc|cc}
        \multicolumn{6}{c}{\textbf{Multi-Step Time Serie Forecasting}} \\ \hline
        \multicolumn{2}{c|}{Aux. Complexity} & \multicolumn{2}{c|}{Emb. Size} & \multicolumn{2}{c}{Emb. Type} \\ \hline
        Constant & Linear & Small & Large & ReLU & LSTM  \\ \hline
        0.958 & 0.953 & 0.962 & 0.958 & 0.958 & 0.950 \\ \hline
    \end{tabular}

    \caption{Ablation study results}
    \label{tab:ablation}
\end{table}

%% file: conclusion.tex
We introduced the SMLE architecture, which augments a backbone network with an embedded, trainable overapproximator, and enables conservative property verification with \emph{controllable complexity}.
We use our architecture to design a framework for training models guaranteed to satisfy desired properties, consisting of two main components: 1) a projection algorithm with delayed constraint generation; and 2) a method to generate strong counterexamples.
We showed how to ground our method on two classes of properties, and demonstrated that it can be competitive with strong baselines, while offering unique advantages, precisely: stronger guarantees compared to preprocessing techniques, and simpler inference compared to postprocessing methods.

Our contributions open up several directions for future research.
First, there are opportunities to reduce the computational cost of our method, e.g. by replacing the first phase of \cref{alg:rt} with traditional adversarial training, by simplifying the counterexample generation problem, or by using a linear cost at projection time.
Second, we believe that our framework might be adapted to address properties involving more than one example, such as fairness constraints, or global monotonicity.
Third, the SMLE architecture itself might be  extended to deal with variable input sizes via Transformers or Graph Neural Networks.

%% file: supplemental_arxiv.tex
\begin{center}
    \Large \textbf{-- Supplemental Material --}
\end{center}

\section{Counterexample Generation for Linear Inequalities}

Let's consider a single linear inequality $R_k \hat{y} \leq r_k$, identified by the $k$-th row in the matrix $R$ and element in the vector $r$.
Restoring feasibility for a counterexample $\clipped{z}$ via translation is in this case a quadratic program with a positive definite matrix.
\begin{equation}
    \delta^* = \argmin_{\delta} \{ \|\delta\|_2^2 \text{ s.t. }
    R_k (\hat{y} + \delta) \leq r_k \}
\end{equation}
where $\hat{y} = \theta_{g,0} + \theta_{g,1:n} \, \clipped{z}$.
As such, its solution can be characterized by means of the Karush Kuhn Tucker (KKT) optimality conditions.
We start by defining the problem Lagrangian:
\begin{equation}
    \mathcal{L}(\delta, \lambda) = \frac{1}{2} \|\delta\|_2^2 + \lambda (R_k (\hat{y} + \delta) - r_k)
\end{equation}
where a scaling factor $\frac{1}{2}$ has been introduced for convenience and
$\lambda$ is the (scalar) multiplier associated to the constraint.
The corresponding KKT conditions are:
\begin{subequations}
\begin{align}
    \delta^T + \lambda R_k & = 0 \\
    R_k (\hat{y} + \delta) & \leq r_k\\
    \lambda_k & \geq 0\\
    \lambda (R_k (\hat{y} + \delta) - r_k) & = 0
\end{align}
\end{subequations}
If $R_k \hat{y} - r_k \leq 0$, meaning that $\hat{y}$ does not represent a counterexample, a valid solution is obtained by setting $\lambda = 0, \delta = 0_v$.
Conversely, if $R_k \hat{y} - r_k > 0$, then we have $\delta = - \lambda R_k^T$ and:
\begin{equation}
    R_k (\hat{y} - \lambda R_k^T) = r_k
\end{equation}
since complementary slackness implies that the constraint $R_k (\hat{y} + \delta) \leq r_k$ will be tight.
From here, we can derive the value of $\lambda$:
\begin{equation}
    \lambda = \frac{1}{R_k^T R_k} (R_k \hat{y} - r_k)
\end{equation}
from which we obtain:
\begin{align}
    \delta = - R_K^T \frac{1}{R_k^T R_k} (R_k \hat{y} - r_k)
\end{align}
which shows that the amount of translation needed to resolve the counterexample is proportional (in terms of absolute value) to the degree of constraint violation.
For simplicity, we drop the constant vector $R_k^T (R_k^TR_k)^{-1}$.
We can account for both considered cases (counterexample/non-counterexample) by using the expression:
\begin{equation}
    \max(0, R_k \hat{y} - r_k)
\end{equation}
By considering every constraint individually in this fashion and summing up, we obtain the formulation presented in the main paper.

\section{Tolerance-Aware Groundings}

When working with practical Mathematical Programming solvers, it is important to account for tolerances to make sure that the models trained with our method actually guarantee property satisfaction.
When grounding our method, this is relevant in two steps, namely when formulating the projection problem and the counterexample generation problem.
When accounting for tolerances, the main goal is \emph{being consistent with the conservative semantic of SMLE verification}.
This can be achieved by:
1) forcing property satisfaction to be slightly too restrictive in the projection problem;
2) allowing for a small degree of constraint violation in the counterexample generation problem.

\paragraph{Linear Inequalities}

We did not encounter issues with tolerances in our experimentation, when dealing with linear properties.
In principle, tolerance-aware formulations for this case can be easily defined.
The one for the projection problem is given by:
\begin{subequations}
\label{eq:tol_prj_lin}
\begin{align}
    \argmin_{\theta_g'} \
    & \|\theta_g' - \theta_g\|_2^2 \\
    \text{s.t. }
    & R \hat{y}_i \leq r - \varepsilon & \forall \clipped{z}_i \in C \label{eq:tol_prj_lin_diff}\\
    \text{with }
    & \hat{y}_i = \theta_{g,0} + \theta_{g,1:n} \, \clipped{z}_i & \forall \clipped{z}_i \in C
\end{align}
\end{subequations}
where $\hat{y}_i$ is actually substituted with $\theta_{g,0} + \theta_{g,1:n} \, \clipped{z}_i$ when building the optimization model, and $\varepsilon$ corresponds to the solver tolerance on constraint satisfaction.
Subtracting the tolerance ensures that, after the projection, all the vectors $\clipped{z}_i \in C$ are no longer counterexamples.

A tolerance-aware formulation for the counterexample generation problem is given by:
\begin{subequations}
\label{eq:tol_ce_lin}
\begin{align}
    \argmax_{x, \clipped{z}} \ & \sum_{k=1}^K U_k (R_k \hat{y} - r_k) \\
    \text{ s.t. }
    & Q(x) \\
    &\clipped{z}_i \geq \underline{h}(x; \theta_{\underline{h}})_i \label{eq:lineq_box_a} \\ 
    \begin{split}
    &\clipped{z}_i \leq B_i \underline{h}(x; \theta_{\underline{h}})_i
    \, + \\
    &(1 - B_i) \bar{h}(x; \theta_{\bar{h}})_i 
    \end{split} & \forall i = 1..n \label{eq:lineq_box_b}\\
    & U_k \in \{0, 1\} & \forall k = 1..K \\
    & B_i \in \{0, 1\} & \forall i = 1..n \\
    &\text{with: } \hat{y}_i = \theta_{g,0} + \theta_{g,1:n} \, \clipped{z} 
\end{align}
\end{subequations}
This is the same formulation reported in the main paper, except that $\hat{y}$ is substituted with $\theta_{g,0} + \theta_{g,1:n} \, \clipped{z}$.
Binary variables $U_k$ and $B_i$ are introduced to model the $\max$ operator, and they can be managed directly by our chosen solver (Gurobi); linearization constraints based on a big-M formulation could be used as an alternative.
Explicitly accounting for constraint satisfaction tolerances in \cref{eq:lineq_box_a,eq:lineq_box_b} is not necessary, since they only \emph{expand} the feasible space: as a result, the solver will consider at least all the counterexamples of the exact formulation.

As we mentioned, when performing our experiments we found that including the tolerance $\varepsilon$ in \cref{eq:tol_prj_lin_diff} was not actually necessary to obtain consistent results.
We conjecture this is due to the fact that, since all relevant constraints are linear, the slight expansion of the feasible space in \cref{eq:tol_ce_lin} is enough to compensate for a small degree of infeasibility in the projection problem.

\paragraph{Mutually Exclusive Classes}

When considering mutually exclusive classes as a property, a tolerance-aware formulation for the projection problem is given by:
\begin{subequations}
\begin{align}
    & \argmin_{\theta_g'}
    \|\theta_g' - \theta_g\|_2^2 \\
    &\quad \text{s.t. }
    I^+_{i,h} + I^+_{i,k} \leq 1 & \forall h, k \in F, \forall \clipped{z}_i \in C \label{eq:mutex_key_cst}\\
    &\quad M I^+_{i,k} - \varepsilon M \geq \hat{y}_{i,k} & \forall k \in O, \forall \clipped{z}_i \in C \\
    &\quad I^+_{i,k} \in \{0, 1\} & \forall k \in O, \forall \clipped{z}_i \in C \\
    &\quad \text{with: } \hat{y}_i = \theta_{g,0} + \theta_{g,1:n} \, \clipped{z}_i & \forall \clipped{z}_i \in C
\end{align}
\end{subequations}
where $\varepsilon > 0$ is the solver constraint and variable tolerance, which we introduced, multiplied by $M$, to ensure that that $\hat{y}_{i,k}$ is assigned with a strictly negative value when the binary $I^+_{i,k}$ is zero, as well as when it is tolerated to be slightly larger than zero.   
Doing this ensures that the semantic of the projection problem is actually conservative.
In our experimentation, such adjustment was crucial to obtain consistent results, since \cref{eq:mutex_key_cst} -- a key constraint for the property formulation -- is stated on variables that are connected to $\hat{y}_i$ via a non-linear expression (i.e. rounding).

A tolerance-aware formulation for the counterexample generation problem is as follows:
\begin{subequations}
\begin{align}
    & \argmax_{x, \clipped{z}}
    I^{m}_{h,k} t_{h,k} \\
    &\quad \text{s.t. }
    Q(x) \\
    &\quad \clipped{z}_i \geq \underline{h}(x; \theta_{\underline{h}})_i \label{eq:lineq_box_a} \\ 
    \begin{split}
    &\quad \clipped{z}_i \leq B_i \underline{h}(x; \theta_{\underline{h}})_i
    \, + \\
    &\quad (1 - B_i) \bar{h}(x; \theta_{\bar{h}})_i 
    \end{split} & \forall i = 1..n \label{eq:lineq_box_b}\\
    &\quad I^m_{h,k} \leq \frac{1}{2}(I^+_{h} + I^+_{k}) & \forall h, k \in F \label{eq:mutex_indicator}\\
    &\quad M - I^+_{k}M \leq \hat{y}_{k} & \forall k \in O \\
    &\quad t_{h,k} \leq \hat{y}_h & \forall h, k \in F \\
    &\quad t_{h,k} \leq \hat{y}_k & \forall h, k \in F\\
    &\quad I^+_{k} \in \{0, 1\} & \forall k \in O \\
    &\quad I^m_{h,k} \in \{0, 1\} & \forall h, k \in F \\
    &\quad \text{with: } \hat{y} = \theta_{g,0} + \theta_{g,1:n} \, \clipped{z}
\end{align}
\end{subequations}
where a fresh variable $t_{h,k}$ is used to linearize the $\min$ operator used in the main paper.
No correction is necessary in this case, except for using a substitution for $\hat{y}$.
In fact, due to the constraint satisfaction tolerance, the solver will treat some pairs $(x, \clipped{z})$ as counterexamples even if this is not actually the case: this behavior is consistent with our conservative semantic and therefore acceptable.

\section{MAP Problems}

The pre- and post-processing approaches used as a baseline for comparison in our experimentation rely on the computation of a constrained Maximum-A-Posteriori (MAP).
In general form, this amounts to solving:
\begin{align}
\label{eq:outputprojection}
\map(y) = \argmax_{y'} \{ \mathcal{L}(y'\mid y) \text{ s.t. } R(y')\}
\end{align}
where $\mathcal{L}(y' \mid y)$ is the likelihood of the adjusted output w.r.t. the reference prediction $y$.
MAP computation problems for specific groundings are then obtained by defining a specific likelihood function to be employed.

For regression tasks, we assume (as commonly done) a Normal distribution and homoskedasticity, i.e. the same variance on all training examples.
Under this assumptions, log-likelihood maximization is equivalent to minimizing the Mean Squared Error.
Accordingly, we use:
\begin{equation}
    \mathcal{L}(y' \mid y) = - \sum_{j=1}^n (y'_j - y_j)^2
\end{equation}
where $n$ is the number of components in the output vector.
For classification tasks, we assume (without loss of generality) a categorical distribution.
Under this assumption, log-likelihood maximization is equivalent to minimizing the cross-entropy.
Accordingly, we use:
\begin{equation}
\begin{split}
    \mathcal{L}(y' \mid y) = - \sum_{j=1}^n
    & (
    y'_j \log \max(\varepsilon, y_j) 
    \, + \\
    & (1- y'_j) \log \max(\varepsilon, 1-y_j)
    )
\end{split}
\end{equation}
where $\max$ operators are introduced to avoid numerical issues in case of values of $y_i$ that are close (or equal) to 0 or 1.
In our experiments, the clipping constant $\varepsilon$ is set to $10^{-6}$.

\section{Experimental Setup}

\paragraph{Hardware Specifications} The software tools used to implement all the considered models, benchmarks and experiments are specified in the main paper. The hardware infrastructure where all the experiments are executed, instead, consists of an Apple M3 Pro CPU with 11 cores and an  Apple M3 Pro GPU with 14 cores, equipped with 36 GB RAM and running macOS v14.1 as operating system.

\paragraph{Benchmarks}
The results presented in the main paper are conducted over three benchmarks: \emph{synthetic regression}, \emph{multi-step time series forecasting} and \emph{multi-label classification}, described in detail as follows.

\textbf{Synthetic Regression}. We design 9 learning tasks, consisting in estimating a vector of the form $F_K(x) = (( \sum_{j=1}^{n} x_j )^k )_{k \in K}$, for input dimensions $n = 2,4,8$ and task difficulties $K = \{1,2\}, \{3,4\}, \{1,2,3,4\}$. For a task $(n, K)$, we enforce 6 linear properties defined as in the main paper, that is, $Qx \leq q \implies Rx \leq r$, where the two systems $Qx \leq q$ and $Rx \leq r$ consist of $m(n) = \log_2(n) + 2$ and $m(K) = \log_2(|K|) + 2$ linear constraints, respectively. The 6 adopted properties, in particular, are selected in order of increasing difficulty (oracle performance) from a pool of similar properties, where each parameter is randomly generated from the interval $[-1,1]$. For each task, we train and test on a dataset of pairs $(x, F_K(x))$, where each component of $x$ is randomly generated from the interval $[-1,1] \setminus S$, with $S$ representing a portion of the main sampling region removed for preliminary out-of-distribution evaluations, whose results are not reported in the paper due to being of marginal interest.
Before executing the actual learning-evaluation process, moreover, the generated data is transformed through common preprocessing routines (e.g. scaling) used in the standard ML workflow; full details on the process can be gained by inspecting the code.

\textbf{Multi-Step Time Series Forecasting}. We consider the problem of estimating the $m$ consecutive values $y = (s_{t+1}, \dots, s_{t+m})$ of a time series, by observing the $n$ previous values $x = (s_{t-n+1}, \dots, s_{t})$ of the same series $S = (s_t)_{t=0}^{t=N}$, where the input and output dimensions $n$ and $m$ are fixed, after a pilot experiment, to $8$ and $4$, respectively. For a series $S$, we consider 3 properties defined as:
$\forall x, |y_i - y_{i+1}| \leq \Delta_q, \forall i=1, \dots, m-1$,
for $q = 0.90, 0.95, 1.00$, where $\Delta_q$ denotes the q-quantile of the set of deviations $\{|s_t - s_{t+1}|\}_{t=0}^{t=N}$.
These restrictions can be encoded as linear properties, and can be thought as a form of stability: we prevent unreasonably high deviations between two consecutive predictions.
We train and test on 25 time series from a public repository \cite{makridakis2020m4}, selected as those on which the \texttt{preprocess} approach guarantees both a good predictive performance and some degree of violation for a least one property $q$. Each series represents a single learning task and is split according to a chronological 80\%-20\% criterion. Moreover, before executing the actual learning-evaluation pipeline, we apply a differencing transformation to the given series $S = (s_t)_{t=0}^{t=N}$, in order to remove trends and seasonality, hence to transform it into a stationary series $\Delta S = (s_{t+1} - s_t)_{t=0}^{t=N-1}$, as common in time series forecasting \cite{box2015time}. Moreover, other than differencing, we preprocess the data by means of standard ML procedures. Full details can be gained by inspecting the code.

\textbf{Multi-Label Classification}. We consider the problem of assigning multiple labels to each instance in a certain dataset $D$. For each $D$, we consider 3 properties of mutually exclusive classes, defined as in the main paper, with $F = \{(c_0, c_1) \in O_D \; |\; \text{freq}_D(c_0, c_1) \leq \Delta_q\}$, where $O_D$ represents the set of possible classes, $\text{freq}_{D}(c_0, c_1)$ the frequency with which the pair $(c_0, c_1)$ occurs among the true labels, while $\Delta_q$ denotes the q-quantile of the set of pair frequencies $\{\text{freq}_{D}(c_0, c_1)\}_{c_0, c_1 \in O_D}$, for $q = 0.0, 0.3, 0.6$. These properties can be thought, again, as a form of stability: we prevent the prediction of unusual combinations.
We train and test on 5 datasets from a public repository \cite{ucodataset}, selected with different cardinalities (average number of labels associated with each instance), roughly ranging from 2 to 26, and different input and output dimensions, ranging from 19 to 103, and from 6 to 14, respectively. Each set represents a single learning task and is split according to a random 80\%-20\% criterion. Moreover, as for the other benchmarks, also in this case we apply standard preprocessing transformations to the data. Full details can be gained, again, by inspecting the code.

The data and code adopted for all three benchmarks are provided in the Code \& Technical Annex. 

\paragraph{Evaluation Metrics}
As evaluation metrics for the predictive performance of the considered models, we adopt the \emph{coefficient of determination} $R^2$ in regression, and the \emph{average class accuracy} AvgAcc in classification. Precisely, given a predictive model $f$ and an evaluation dataset $D$, they are defined as:

\begin{subequations}
\begin{align}
    R^2 & = 1 - \frac{\sum_{(x,y) \in D} (y - f(x))^2}{\sum_{(x,y) \in D} (y - \bar{y})^2}  \label{eq:evaluation_metrics_reg}\\
    \text{AvgAcc} & = \frac{1}{m} \sum_{j=1}^{m} \left( \frac{1}{|D|} \sum_{(x, y) \in D} \mathbf{1}(y_j = f(x)_j) \right) \label{eq:evaluation_metrics_class}
\end{align}
\end{subequations}
with $\bar{y} = \sum_{(x,y) \in D} y$ in \cref{eq:evaluation_metrics_reg} denoting the mean of the ground-truth labels, while $m$ in \cref{eq:evaluation_metrics_class} the total number of classes (i.e., the output dimension).

These metrics represent two commonly adopted evaluation criteria for ML models, thanks to their simplicity and interpretability. Moreover, when used together, they allow the evaluation and comparison of both regressors and classifiers within a common scale, given that they both consist of a value ranging from a minimum of 0 to a maximum of 1, which we find particularly convenient in our case.

\input{tables/architectures}

\input{tables/architectures_byexp}

\paragraph{Model Parametrization}
The main model hyperparameters in our computational study are the architectures adopted for $f = g \circ h$, affecting \texttt{preprocess}, \texttt{postprocess} and  \texttt{smle}, and for the auxiliary models $\underline{h}$ and $\bar{h}$, impacting instead solely \texttt{smle}. These key design choices are tailored to the different research questions and benchmarks. In particular, the entire set of architectures used in our experiments is displayed in \Cref{tab:architectures}, where $\sigma_k$ denotes a network layer with $k$ nodes and $\sigma$ as activation function, while $n$ and $m$ represent the input and output dimensions, respectively. \Cref{tab:architectures_byexp} shows instead how these architectures are combined in each section of the computational study reported in the main paper. Moreover, as in any standard Deep Learning setting, our study also requires the selection of the hyperparameters controlling the training processes. Their configuration is detailed, for each benchmark, in \Cref{tab:training_hyperparameters}.

In each experiment, every competitor is trained once for each task and property. Note moreover that, while our framework \texttt{smle} is trained, by default, on property-enforced data via the use of the MAP operator (\cref{eq:outputprojection}), as for \texttt{preprocess}, in the ablation study conducted over the Synthetic setup, reported in Q3, we train instead on the original data. This decision stems from preliminary and less relevant experiments, not reported in the paper.

The hyperparameters listed in \Cref{tab:architectures_byexp,tab:training_hyperparameters} were determined based on pilot experiments aiming at maximizing predictive performance, or inherited from the default setting adopted by the libraries used for the implementation. 

Finally, beyond the model and training hyperparameters, our computational study involves an additional set of minor, less critical decisions (e.g. random seeds for randomness-affected routines, training initialization, etc), which can be obtained by inspecting the code provided in the Code \& Data Annex.  

\input{tables/training_hyperparameters}

%% file: tables/architectures.tex
\begin{table}[h!]
    \centering
    \renewcommand{\arraystretch}{1.5}
\begin{tabular}{p{0.08cm}|p{7.47cm}}
    \hline
    \multirow{2}{*}{$g$} & $g_1 : \text{linear}_m$ \\ 
    & $g_2 : \text{linear}_{\left\lfloor 4\log_2(nm) \right\rfloor}$ \\ \hline
    \multirow{4}{*}{$h$} & $h_1 : \text{relu}_{nm} \circ \text{relu}_{2nm} \circ \text{relu}_{nm}$ \\
    & $h_2 : \text{relu}_{nm} \circ \text{relu}_{2nm} \circ \text{relu}_{3nm} \circ \text{relu}_{2nm} \circ \text{relu}_{nm}$ \\ 
    & $h_3 : \text{lstm}_{nm} \circ \text{relu}_{2nm} \circ \text{relu}_{3nm} \circ \text{relu}_{2nm} \circ \text{relu}_{nm}$ \\ 
    & $h_4 : \text{relu}_{\left\lfloor 4\log_2(nm) \right\rfloor} \circ \text{relu}_{\left\lfloor 8\log_2(nm) \right\rfloor} \circ \text{relu}_{\left\lfloor 4\log_2(nm) \right\rfloor}$ \\ \hline
    \multirow{3}{*}{\makecell{$\underline{h}$ \\[3pt] $\bar{h}$}} & $\underline{h}_1, \bar{h}_1 : \text{linear}_{nm}$ \\
    & $\underline{h}_2, \bar{h}_2 : \text{constant}_{nm} $ \\
    & $\underline{h}_3, \bar{h}_3 : \text{constant}_{\left\lfloor 4\log_2(nm) \right\rfloor}$ \\
    \hline
\end{tabular}
\caption{Neural Network Architectures}
\label{tab:architectures}
\end{table}

%% file: tables/architectures_byexp.tex
\begin{table}[h!]
    \centering
    \renewcommand{\arraystretch}{1.5}
\begin{tabular}{c|ccc}
    \multicolumn{4}{c}{\textbf{Q1\&Q2}} \\ 
    \cline{2-4}
    \multicolumn{1}{c}{} & Synthetic & Forecasting & Classification \\ \hline
    $g$ & $g_1$ & $g_1$ & $g_2$ \\
    $h$ & $h_1$ & $h_2$ & $h_4$ \\
    $\underline{h}, \bar{h}$ & $\underline{h}_1, \bar{h}_1$ & $\underline{h}_2, \bar{h}_2$ & $\underline{h}_3, \bar{h}_3$ \\ \hline
\end{tabular}

\vspace{3pt}

\begin{tabular}{c|c|c|c}
    \multicolumn{2}{c}{} & \multicolumn{2}{c}{\textbf{Q3}} \\
    \cline{3-4}
    \multicolumn{2}{c}{} & Synthetic & Forecasting \\ \hline
    \multirow{3}{*}{\rotatebox{90}{Aux. Comp.}}
    & $g$ & $g_1$ & $g_1$ \\ 
    & $h$ & $h_1$ & $h_2$ \\
    & $\underline{h}, \bar{h}$ & $\underline{h}_1, \bar{h}_1$ vs $\underline{h}_2, \bar{h}_2$ & $\underline{h}_1, \bar{h}_1$ vs $\underline{h}_2, \bar{h}_2$ \\ \hline
    
    \multirow{3}{*}{\rotatebox{90}{Emb. Size}}
    & $g$ & $g_1$ & $g_1$ \\ 
    & $h$ & $h_1$ vs $h_2$ & $h_1$ vs $h_2$ \\
    & $\underline{h}, \bar{h}$ & $\underline{h}_1, \bar{h}_1$ & $\underline{h}_2, \bar{h}_2$ \\ \hline
    
    \multirow{3}{*}{\rotatebox{90}{Emb. Type}}
    & $g$ & -- & $g_1$ \\ 
    & $h$ & -- & $h_2$ vs $h_3$ \\
 & $\underline{h}, \bar{h}$ & -- & $\underline{h}_2, \bar{h}_2$ \\ \hline
\end{tabular}
\caption{Architecture configuration for each experiment}
\label{tab:architectures_byexp}
\end{table}

%% file: tables/training_hyperparameters.tex
\begin{table}[t]
    \centering
    \renewcommand{\arraystretch}{1.}

    \begin{tabular}{r|ccc}
        \multicolumn{1}{c}{} & Synthetic & Forecasting & Classification \\ \hline 
         val. split & 0.2 & 0.2 & 0.2 \\
         optimizer & adam & adam & adam \\  
         loss & mse & mse & bin. crossent. \\
         batch size & 128 & 32 & 32 \\
         max. epochs & 1000 & 1000 & 1000 \\
         stop. patience & 5 & 15 & 30 
    \end{tabular}
    
    \caption{Training hyperparameters}
    \label{tab:training_hyperparameters}
\end{table}